\documentclass[10pt,twocolumn,letterpaper]{article}

\usepackage{iccv}
\usepackage{times}
\usepackage{epsfig}
\usepackage{graphicx}
\usepackage{amsmath}
\usepackage{amssymb}
\usepackage{bm}

\newcommand{\mypar}[1]{\vspace{2mm}\noindent\textbf{#1}}
\newcommand{\thetab}{\bm{\theta}}
\newcommand{\betab}{\bm{\beta}}
\newcommand{\Tb}{\mathbf{T}}

\newcommand{\methodN}{HUSC-S }
\newcommand{\methodFull}{HUSC }

\iccvfinalcopy 

\def\iccvPaperID{5873} 

\ificcvfinal\fi
\begin{document}

\title{Human Synthesis and Scene Compositing}

\author{Mihai Zanfir$^2$ \hspace{2mm}  Elisabeta Oneata$^2$ \hspace{2mm}  Alin-Ionut Popa$^2$   \hspace{2mm}    Andrei Zanfir$^2$  \hspace{2mm}     Cristian Sminchisescu$^{1, 2}$\\{\tt\small \{mihai.zanfir, elisabeta.oneata, alin.popa, andrei.zanfir\}@imar.ro,}
\\{ \tt\small cristian.sminchisescu@math.lth.se} \\ 
$^1$Department of Mathematics, Faculty of Engineering, Lund University\\ $^2$Institute of Mathematics of the Romanian Academy}

\makeatletter

\makeatother

\maketitle
\begin{abstract}
    Generating good quality and geometrically plausible synthetic images of humans with the ability to control appearance, pose and shape parameters, has become increasingly important for a variety of tasks ranging from photo editing, fashion virtual try-on, to special effects and image compression. 
    In this paper, we propose a \methodFull (\textbf{HU}man \textbf{S}ynthesis and Scene \textbf{C}ompositing) framework for the realistic synthesis of humans with different appearance, in novel poses and scenes. Central to our formulation is 3d reasoning for both people and scenes, in order to produce realistic collages, by correctly modeling perspective effects and occlusion, by taking into account scene semantics and by adequately handling relative scales. Conceptually our framework consists of three components: (1) a human image synthesis model with controllable pose and appearance, based on a parametric representation, (2) a person insertion procedure that leverages the geometry and semantics of the 3d scene, and (3) an appearance compositing process to create a seamless blending between the colors of the scene and the generated human image, and avoid visual artifacts.
    The performance of our framework is supported by both qualitative and quantitative results, in particular state-of-the art synthesis scores for the DeepFashion dataset. 
\end{abstract}

\section{Introduction}

Generating photorealistic synthetic images of humans, with the ability to control their shape and pose parameters, and the scene background is of great importance for end-user applications such as in photo-editing or fashion virtual try-on, and for a variety of data-hungry human sensing tasks, where accurate ground truth would be very difficult if not impossible to obtain (\eg the 3d pose and shape of a dressed person photographed outdoors). 

One way to approach the problem would be to design human models and 3d environments using computer graphics. While the degree of realism increased dramatically in narrow domains like the movie industry, with results that pass the visual Turing test, such synthetic graphics productions requires considerable amount of highly qualified manual work, are expensive, and consequently do not scale. In contrast, other approaches avoid the 3d modeling pipeline altogether, aiming to achieve realism by directly manipulating images and by training using large-scale datasets. While this is cheap and attractive, offering the advantage of producing outputs with close to real statistics, they are not nearly as controllable as the 3d graphics ones, and results can be geometrically inconsistent and often unpredictable. 

\begin{figure}[!htbp]
\begin{center}
         \includegraphics[width=1\linewidth]{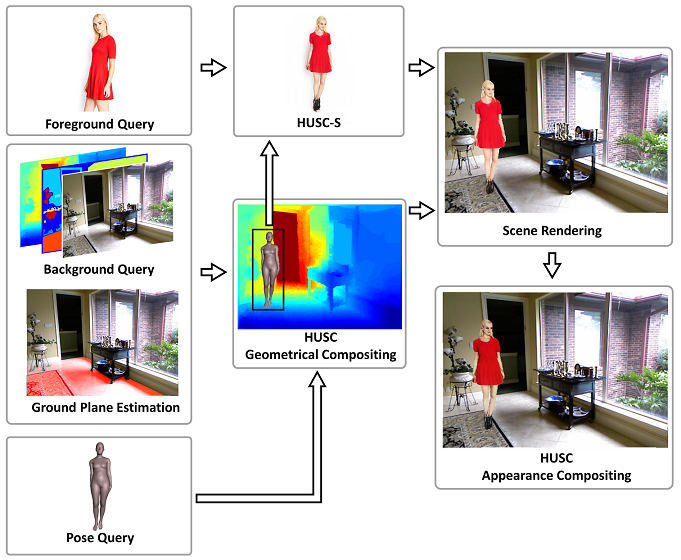}
\end{center}

\caption{Overview of our full pipeline HUSC. Inputs are a foreground image of a person, a background scene with its associated depth map and semantic labeling, and a target 3d body model. First, we perform ground plane estimation of the background scene. Within the \textbf{Geometrical Compositing} stage, we sample a valid 3d location for the target body and perform the associated viewpoint transformation and alignment with the supporting plane normal. The newly updated target body shape, together with the input image encoding the desired appearance, are passed to the human synthesis network, HUSC-S. The resulting synthesized foreground image is rendered in the background scene, by properly accounting for depth ordering constraints. Finally, its appearance is altered by our learned \textbf{Appearance Compositing} network in order to produce the final result.}
\label{fig:HUCS}
\end{figure}

In this work we attempt to combine the relatively accessible methodology in both domains and propose a framework that is able to realistically synthesize a photograph of a person, in any given pose and shape, and blend it veridically with a new scene, while obeying 3d geometry and appearance statistics. An overview is given in fig. \ref{fig:HUCS}.
For the first part of human synthesis, given a source image of a person and a different target pose (or more generally, a target 3d body mesh), we want to generate a realistic image of the person synthesized into the new pose. We would like that all the elements included in the person's source layout (either clothing, accessories or body parts) to be preserved or plausibly extended in the synthesis. 

We propose to learn a dense displacement field, that leverages 3d geometry and semantic segmentation (\eg a blouse \textit{moves} differently than a hat; the leg \textit{moves} differently than the head). This module produces a correction to an initial body displacement field and is trained jointly with the synthesis model within a single end-to-end architecture. 

Finally, given an image of a person (real or synthesized) and of a background scene, we want to generate a good quality composite of the two. We argue that a realistic synthesis should consider the physical properties of the scene and of the human body, and also compensate for the different appearance statistics. With that in mind, we propose to blend the foreground image with the background image, at two levels: geometry and appearance. At the geometric level, we want the foreground person, with its associated 3d body model, to respect the 3d space and scale constraints of the scene, be visible according to the scene depth ordering, and be placed on a plausible support surface (\eg floor). At the appearance level, we would like that the two sources blend naturally together, without the undesirable cut-and-paste look. To summarize, our contributions are as follows: \emph{(a)} a realistic human appearance translation task, with state-of-the-art results, \emph{(b)} a realistic data augmentation procedure, which allows for the synthesis of complex scenes containing humans, with available pseudo-ground-truth labels such as: pose, shape, segmentation and depth.

\begin{figure*}[!htbp]
\begin{center}
        \scalebox{0.85}{
         \includegraphics[width=1\linewidth]{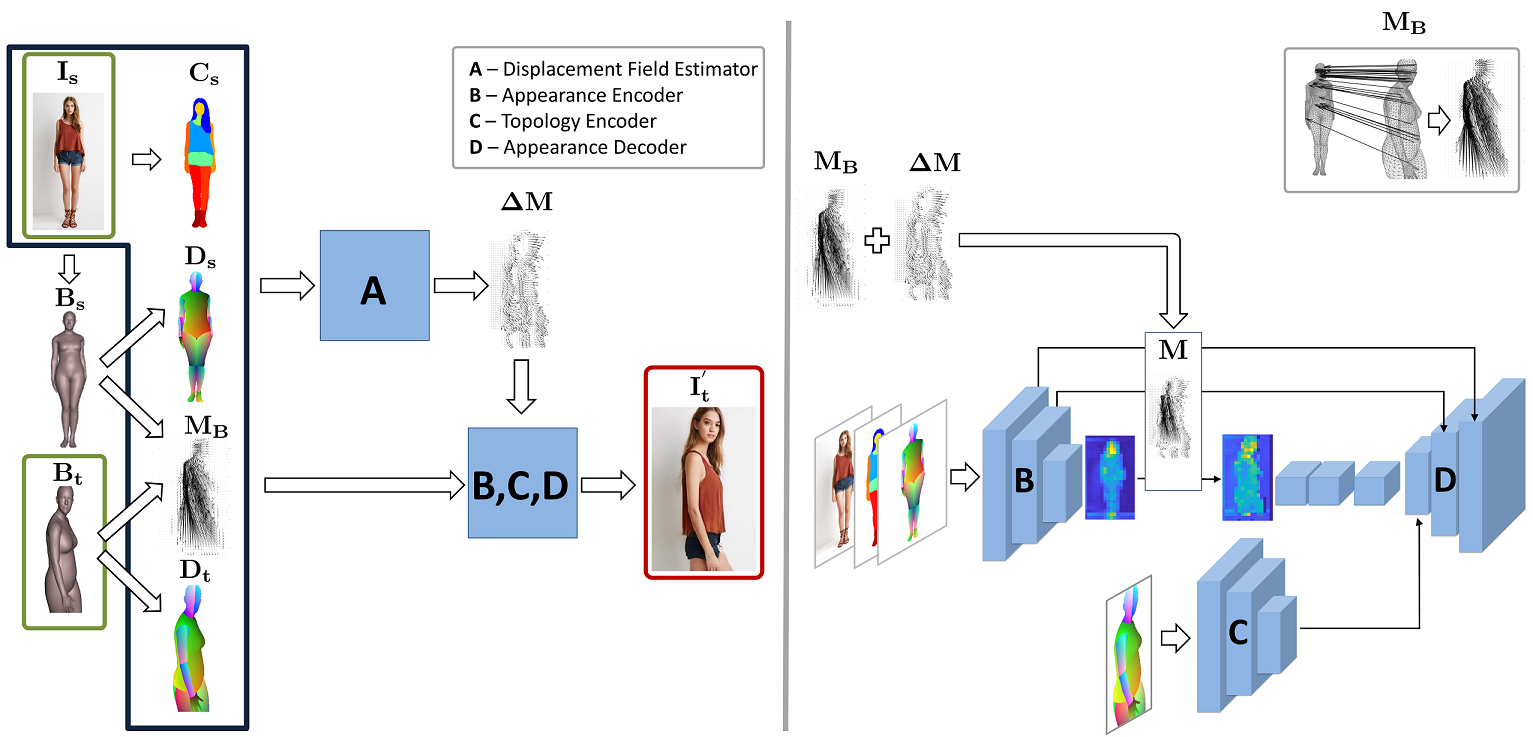}}
\end{center}
\caption{ \textbf{(Left)} Overview of HUSC-S. Our pipeline receives as input (shown inside green boxes) a source monocular image of a person, $I_s$, and a desired target 3d body mesh, $B_t$. We estimate for $I_s$ a 3d body mesh, $B_s$, and a clothing segmentation, $C_s$. From $B_s$ and $B_t$, we can compute a dense body displacement field $M_B$, and also their respective dense pose representations, $D_s$ and $D_t$. Our network receives as input to \textbf{displacement field estimator (A)} on the top branch, all the available information, for both source and target. It outputs an update $\Delta M$ that is added to $M_B$ to produce the final displacement field, $M$. \textbf{(Right)} Detailed view of HUSC-S. The  \textbf{appearance encoder (B)} receives as input only the information pertaining to the source. It produces features maps at different resolutions that are all displaced according to $M$. We show an example of such a transformation at the lowest resolution. The \textbf{topology encoder (C)} operates on the target dense pose, $D_t$, and is input to the decoder \textbf{D} to help guide the synthesis, alongside all the feature maps coming from \textbf{B}. The output of our network is the final synthesized image $I_{t}^{'}$ (shown inside the red box, left).}
\label{fig:HS_overview}
\end{figure*}

\section{Related Work}
\mypar{Human Image Synthesis.} 
An important body of work in the literature is dedicated to image synthesis \cite{goodfellow2014,nguyen2016plug,isola2017,yi2017dualgan,DRIT,wang2018pix2pixHD,zhang2018residual,luan2017deep,Johnson2016Perceptual}, and more specifically to the task of synthesizing photo-realistic images of humans
\cite{zanfir18human,ma2017pose,han2018viton,ma2018disentangled,siarohin2018deformable,lassnergenerative, Esser_2018_CVPR,Soft-Gated_nips18,Grigorev2018CoordinatebasedTI}. Among these, a significant proportion  -- our work included -- have focused on synthesizing humans given a condition image and a desired target pose. In \cite{Esser_2018_CVPR} the authors propose a variational U-Net for conditional image generation. Their method synthesizes images of humans based on 2d pose information and a latent appearance representation learned with a variational auto-encoder. We leverage a richer shape and position representation in the form of a 3d body model and learn a dense correspondence field between the pose and shape of the source person and that of the target. This motion field is extended beyond body regions to clothing and hair. \cite{Soft-Gated_nips18} learns a transformation grid based on affine and thin-plate spline embedding, which is used to warp the condition image in the desired target position.

In contrast, our learned displacement field allows for dense arbitrary transformations. Our work relates to \cite{siarohin2018deformable}, through the use of deformable skip connections for warping the features maps of the conditioning image, in the location of the desired target pose. However, there are two major differences between our method and the DSCF net proposed in \cite{siarohin2018deformable}: i) In DSCF net, the image features are warped with an affine transformation obtained through an optimization step both at training and at testing time. 

ii) The deformable skip connection in DSCF net transform feature-maps, that correspond to coarse 2d body joint activations, while our learned displacement field is densely computed over the entire person layout (body and clothing). Different from \cite{zanfir18human}(HAT), we learn end-to-end the dense correspondence field, coupling it together with the synthesis part of our network. HAT operates in two different stages: one that synthesizes color only on the target body shape, and one that tries to further complete the color on the outside regions. This latter stage is completely detached from the initial source image and is guided only by a target clothing segmentation.

\mypar{General data augmentation through image synthesis.} Combining different sources of synthetic data with real data, and then generating a realistic composition of the both, is an increasing popular research direction. It has been successfully applied to various tasks, such as semi-supervised foreground-background segmentation \cite{remez2018learning, alhaija20018geometric, dwibedi2017cut}, object detection \cite{dvornik2018modeling, dwibedi2017cut} or 3d object pose estimation \cite{alhaija20018geometric}. Different than \cite{alhaija20018geometric}, we do not have access to a realistic renderer that -- in our case -- produces humans with various garments and different poses. That, in itself, represents another complex research topic. The two cut-and-paste methods \cite{dwibedi2017cut, remez2018learning} use simple blending techniques, and only for the foreground object, while we propose to learn a blending for both the background and foreground, accordingly. Note that while \cite{alhaija20018geometric} and \cite{dwibedi2017cut} take the 3d geometry of the scene into account, they only consider 3d rigid objects.

\mypar{Human Synthesis on Backgrounds.}  Most of the previous works focus solely on the human appearance, and are not concerned with a natural blending within a background image. There are some works that consider synthesizing the entire scene, such as \cite{varol17_surreal}, where textured human body models are overlaid over random backgrounds. This method does not consider the semantic and the 3d structure of the scene, hence, the human model is often placed in unnatural locations. Moreover, the model's dimensions and illumination do not match the ones of the surrounding environment. Another relevant work is the one of  \cite{balakrishnan2018synthesizing}, that takes as input an image containing a human and its desired 2d skeleton configuration. The method changes the position of the human to the given pose configuration, while retaining the background scene and inpainting the missing areas. 

\mypar{Image Adjustment.} 

Our work on image compositing is inspired by \cite{tsai2017deep}, where given an initial composite image and a foreground mask, a deep network learns to adjust the foreground region such that it becomes compatible (harmonized) with the background region. While their method is applied for arbitrary foreground objects, we focus on the specific case of compositing humans into backgrounds. We use a different type of encoder-decoder architecture with ResNet blocks inspired by \cite {wang2018pix2pixHD}. Moreover, our method does not need an extra task of semantic scene parsing in decoding the harmonized image.

\section{Human Synthesis}
\mypar{Overview.} Given a source RGB image $I_s$ of a clothed person, our goal is to realistically synthesize the same person in a given novel pose, in our case represented by a 3d body model, $B_t$. Different from previous works that rely mostly on 2d joint to model the pose transformation, one of our key contributions lies in incorporating richer representations, such as 3d body shape and pose which improves the predictions in complicated cases -- e.g. determining how the position/occlusion of the body and clothing/hair change with the motion/articulation of the person. Moreover, a 3d body model is required to plausibly place the person in the scene and allows us to  have full control over 3d body shape, pose, and the scene location where the synthesized person is inserted.

Our human synthesis model is a conditional - GAN \cite{wang2018pix2pixHD} that consists of a generator $\mathcal{G}$ and a discriminator $\mathcal{D}$. Given a source image $I_s$ and a condition pose $B_t$, the task of the generator is to produce an output image $I_t' = \mathcal{G} (I_s)$ of the person in  $I_s$  having the body pose $B_t$.  The discriminator's objective is to distinguish between real and generated images while the generator's task is to produce fake images which are able to fool the discriminator.

Starting from an encoder-decoder architecture, our key contribution is the addition of a novel \textit{Displacement Field Estimator} in the generator, that learns to displace features from the encoder, prior to being decoded. Our modified architecture also relies on the semantic segmentation (i.e. clothing and body parts) of the source image, $C_s$,
and on an estimated 3d body model of the source, $B_s$. An overview of our generator can be seen in fig. \ref{fig:HS_overview}.  We use a multiscale discriminator as in \cite{wang2018pix2pixHD}.

\mypar{3d Body Model Estimation.} For estimating the 3d body model of an image, we use the SMPL model \cite{SMPL2015} and follow a similar procedure to \cite{zanfir17}, but instead apply the method of \cite{zanfir_nips2018} for 3d pose estimation. The SMPL model is a parametric 3d human mesh, controlled by pose parameters $\thetab \in \mathbb{R}^{24\times 3}$ and shape parameters $\betab \in \mathbb{R}^{10}$, that generate vertices $\mathbf{V}(\thetab, \betab) \in \mathbb{R}^{6890\times3}$. We fit this model to predicted 2d keypoints and semantic segmentation, under a full-perspective model with fixed camera intrinsics. We refer to the tuple $(\thetab, \betab, \Tb)$ as a model $B$, where $\Tb$ is the inferred camera-space translation.

\mypar{Dense Body Correspondences.} Given two body meshes (i.e. a source and a target) we can compute a 3d displacement field $M_B^{3d}$ of the visible surface points on the source $\mathbf{V_s}(\thetab_s, \betab_s)$, to the target surface points  $\mathbf{V_t}(\thetab_t, \betab_t)$ (see fig. \ref{fig:HS_overview}, top right corner). 
\begin{align}
    M_B^{3d} = \mathbf{V_s}(\thetab_s, \betab_s) - \mathbf{V_t}(\thetab_t, \betab_t)
\end{align}

The displacement field is projected in 2d, and encoded onto the target shape, representing the offset from where information is being transferred. We will refer to this final 2d displacement field as $M_B = \mathcal{P}(M_B^{3d})$.

\mypar{Topological Representation.} From a 3d body mesh we obtain a 2d dense pose \cite{Güler_2018_CVPR} representation for both the source, $D_s$, and the target, $D_t$. The dense pose representation encodes at pixel level, for the visible 3d body points, the semantic body part it corresponds to, alongside the 2d coordinates in the local body part reference system. Each 3d body point will therefore have a unique associated value using this encoding scheme.

\mypar{Image Synthesis Network.} Our image synthesis network generator (HUSC-S) has two novel computational units: a \textit{Displacement Field Estimator}, and a \textit{Topology Encoder}. 
 The former learns an update, $\Delta M$, for the displacement field $M_B$, which is meant to either correct the warping induced by an erroneous fitting, or capture motions that fall outside the fitted model (e.g. hair, clothing). The final displacement field is given by $M=\Delta M + M_B$, and will be used to place source features at the correct spatial location in the target image, to facilitate decoding and thus, synthesis. This update step is computed on geometric and semantic features, as follows: $I_s$, $D_s$, $D_t$, $C_s$ and $M_B$.
 
An \textit{Appearance Encoder} module extracts features $A_s$ based on the RGB image $I_s$, a pre-computed clothing segmentation $C_s$ (using \cite{Gong_2018_ECCV}) and the source dense pose, $D_s$. The updated displacement field $M$ is used to warp the appearance features $A_s$ and produce warped image features $A_t^{'}$. These are then passed through a series of residual blocks. The result is concatenated with an encoding of $D_t$, produced by the \textit{Topology Encoder}, and then passed to an \textit{Appearance Decoder} to generate the synthesized image, $I_t^{'}$. The addition of the encoded $D_t$ is useful to the decoding layers, because the underlying body-part topology, on which they operate, is exposed to them. We add deformable (i.e. warped by our estimated displacement field) skip connections between the \textit{Appearance Encoder} and the \textit{Appearance Decoder}, to propagate lower level features directly to the decoding phase.

\mypar{Training.} At training time, we are given pairs of images ${\{(I_{s}, I_{t})\}}$ which are used to learn both the parameters of the discriminator, $\theta_d$, and those of the generator, $\theta_g$, using a combined $L1$ color loss, a perceptual VGG loss \cite{Johnson2016Perceptual} and a discriminator GAN loss.
\begin{align}
\mathcal{L} = \mathcal{L}_{GAN} + \lambda\mathcal{L} _{L1} + \gamma\mathcal{L}_{VGG}
\end{align}

\begin{align}
\mathcal{L}_{GAN} = \mathbb{E}_{(I_s, I_t)}[\log \mathcal{D}(I_s, I_t)] + \mathbb{E}_{I_s}[\log (1 - \mathcal{D}(I_s, \mathcal{G}(I_s))]
\end{align} 

\begin{align}
\mathcal{L}_{L1} = \sum_{(u, v)} \left | \left |I_t(u, v) - I_t'(u, v)  \right| \right |_1
\end{align} 

\begin{align}
\mathcal{L}_{VGG} = \sum_{i = 1}^{N}\frac{1}{N_i}\left|\left| VGG^{(i)} (I_t) - VGG^{(i)}(I_t') \right | \right |_1
\end{align} 

$N$ represents the number of layers considered from the VGG network,  $N_i$ is the dimension of the output features and $VGG^{(i)}(I) $ are the activations of the $i^{th}$ layer on  image $I$.

\section{Human and Scene Compositing}

In image compositing, we are interested in combining two visual elements from different sources into a single image, creating the impression that they are actually part of the same scene. In order to create this effect, the addition of the person in the scene should respect the physical laws of the real world, and the images statistics of the person (given by illumination, contrast, saturation, resolution) should match those of the scene. To meet our goals, we consider that we have a background image with associated depth and semantic information, and a foreground image of a person (synthesized or real) with an estimated 3d body model (see fig. \ref{fig:HUCS} for an overview).

\subsection{Geometrical Compositing} 

We are interested in generating images that respect real world physical constraints, such as: people are (usually) sitting on a ground plane, do not penetrate objects in the scene and have plausible body proportions considering the scale of the environment. In order to estimate a ground plane, we select all 3d points that correspond to the semantic classes associated with a floor plane surface. We fit a plane using a least-squares solution, combined with a RANSAC approach to ensure robustness to outliers. This is necessary to mitigate the noise in the depth signal.

In order to choose a physically plausible position for the body model in the scene, we sample 3d floor locations on a regular grid. To make the model perpendicular to the plane, we compute a rotation matrix that aligns the "up" vector of the model to the plane normal, and multiply it with the original rotation matrix. Also, due the computed camera-space translation, the relative camera orientation will change. Adjusting for both the rotation induced by the plane normal alignment, and the camera point-of-view, involves a rotation of the original SMPL model. To be able to correct the appearance change, associated with this transformation, we run our HUSC-S component for the foreground image and its newly created target 3d model. Notice that this is not at the reach of other methods, as they do not have the means to correct their foreground hypotheses based on 3d transformations.

Finally, we test if the final mesh model collides with other 3d points in the scene. We approximate the body model with simple 3d primitives and check for intersections. If no collisions are detected, we can then place the foreground image. We update the color of a pixel according to the relative ordering of points in depth, belonging to both the model and the scene (e.g if the model is behind a desk, then some of its body parts will be occluded). 

One important property of this procedure is its scalability: none of the components requires human intervention, as the whole pipeline is completely automated. The Inception Score \cite{salimans2016improved}, or other metrics, can be used to detect and prune the possible poor quality generated images.

\subsection{Appearance Compositing}\label{sec:appearance_compositing}

\begin{figure}[!htbp]
\begin{center}
         \includegraphics[width=0.99\linewidth]{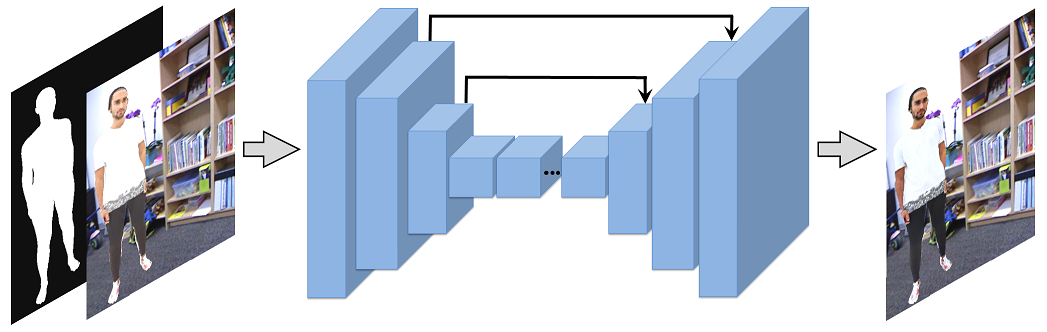} 
\end{center}
\caption{Our Appearance Compositing architecture is represented by an encoder-decoder network inspired by the generator used in \cite{wang2018pix2pixHD}.  
It takes as input the geometrical composited imaged and a figure ground segmentation of the person, and corrects his appearance in order to naturally match the scene of insertion. }
\label{fig:AppCompositing}
\end{figure}

In order to create a natural composition of a foreground and background image pairs, we build on a methodology originally proposed in  \cite{tsai2017deep} for generating realistic composites of arbitrary object categories over backgrounds. (see fig. \ref{fig:AppCompositing} for an overview). We learn an adjustment of the color and boundaries of a human composite such that it blends naturally (in a statistical, learned sense) in the scene. For training we use images with humans from COCO. We alter image statistics inside the human silhouette so that it looks unnatural, thus simulating a silhouette pasted on a different background. The network learns to recreate the original image.

We make a series of adjustments to the methodology proposed in \cite{tsai2017deep}. First, we restrict the problem to generating realistic composites of only humans and backgrounds. We build our model based on a different, more recent network architecture \cite{wang2018pix2pixHD} and we drop the Scene Parsing task since our compositing problem is defined on a specific category (humans).  The network's input is a concatenation of the modified image and the binary figure-ground mask ( see fig. \ref{fig:AppCompositing}). We add skip connections between the encoder and decoder, such that the network can easily access background features in the decoding stage. We also remove the batch normalization layers, so that the original image statistics are preserved. We train the network by using a combined $L1$ color loss, a perceptual VGG loss \cite{Johnson2016Perceptual} and a discriminator GAN loss.

\section{Experimental Results}

\mypar{Evaluation metrics.} To test the quality of our generated images we use several metrics: the Learned Perceptual Image Patch Similarity metric (LPIPS) \cite{zhang2018perceptual}, the Inception Score (IS) \cite{salimans2016improved} and the Structural Similarity Index (SSIM) \cite{Wang04imagequality}. The LPIPS metric was learned using a database of images on which different distortions were applied. 
IS provides a measure of how realistic a generated image is. This is obtained by learning a correlation with human judgment over synthetic versus real images. However, according to \cite{neverova2018dense}, the IS score of the images present in the DeepFashion dataset is already very low (an average of $3.9$ vs an average of $11.2$ for CIFAR-10 images), which makes this metric not conclusive with respect to the level of realism of the generated samples. The SSIM score measures the level of structural degradation in the synthesized image, compared to the reference image.

\begin{figure}[!htbp]
\begin{center}
\scalebox{0.93}{
\begin{tabular}{ccc}
       \includegraphics[width=0.99\linewidth]{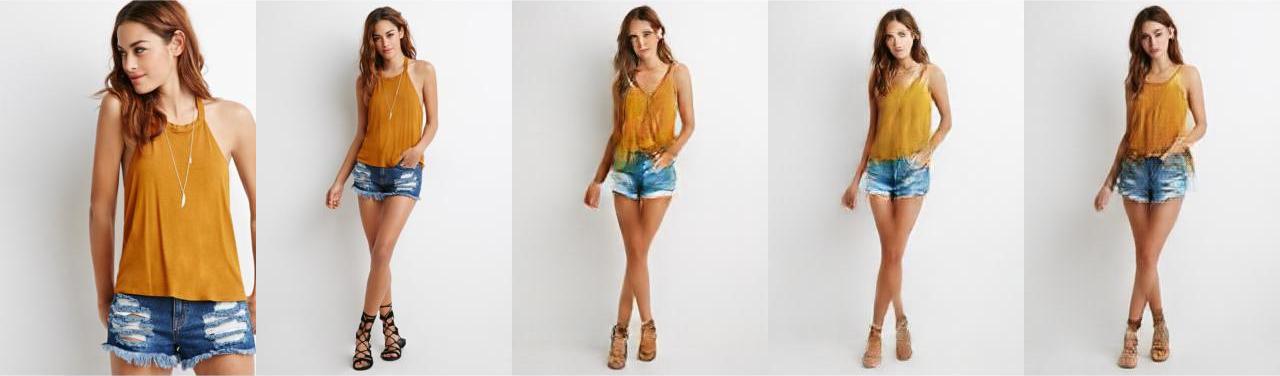}    \\
        \includegraphics[width=0.99\linewidth]{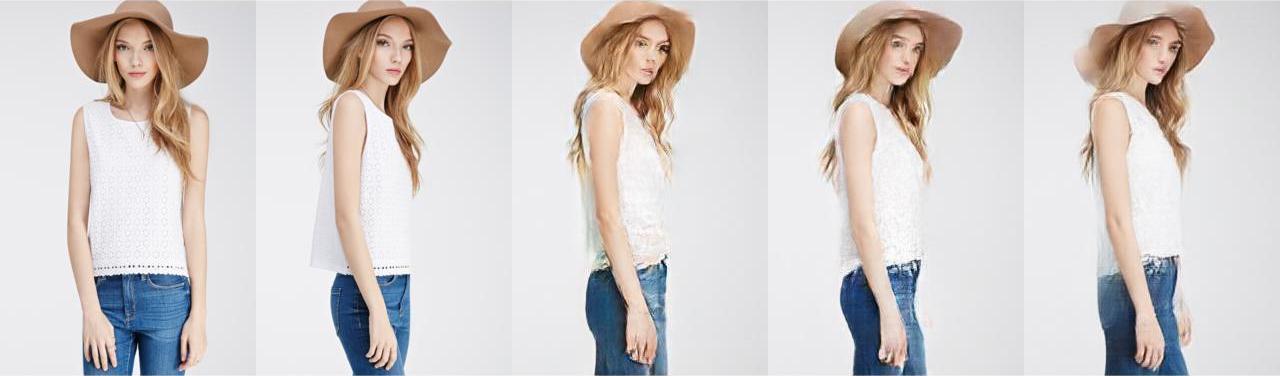}\\
        \includegraphics[width=0.99\linewidth]{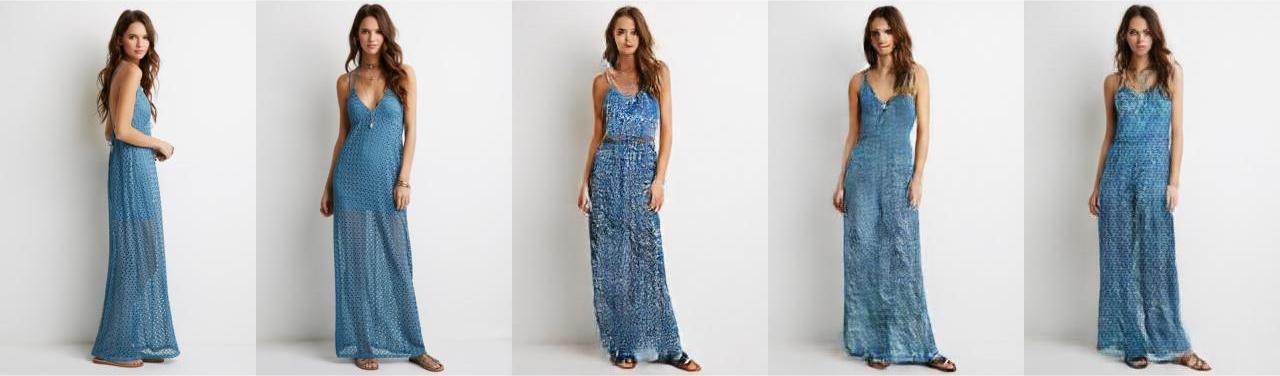}\\
        \includegraphics[width=0.99\linewidth]{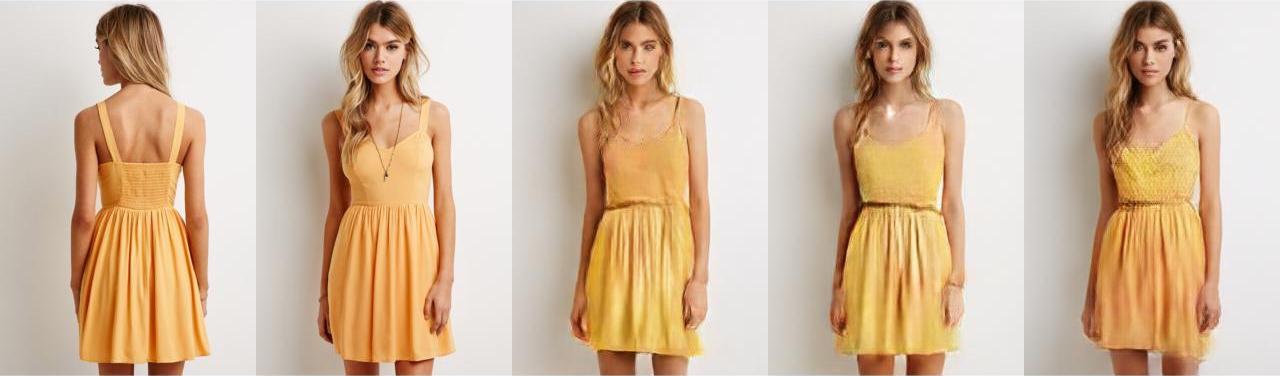}\\
        \includegraphics[width=0.99\linewidth]{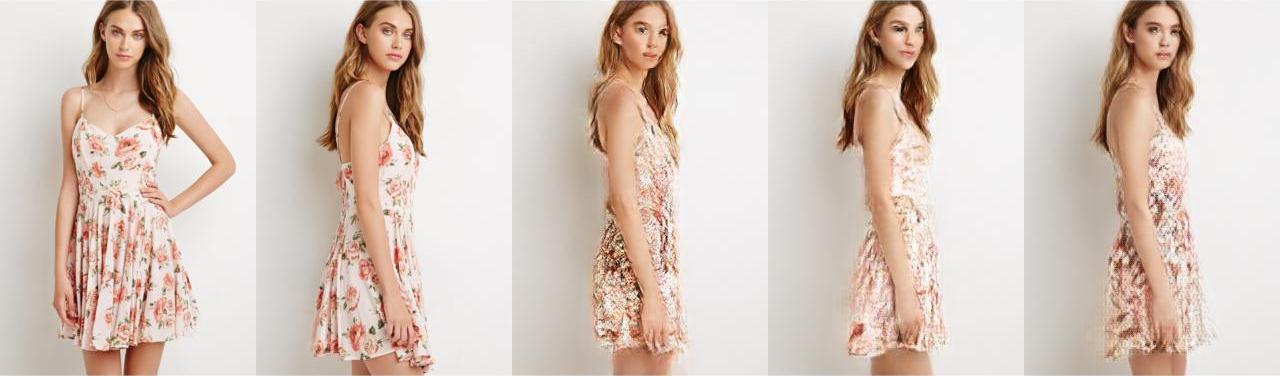}\\
         \includegraphics[width=0.99\linewidth]{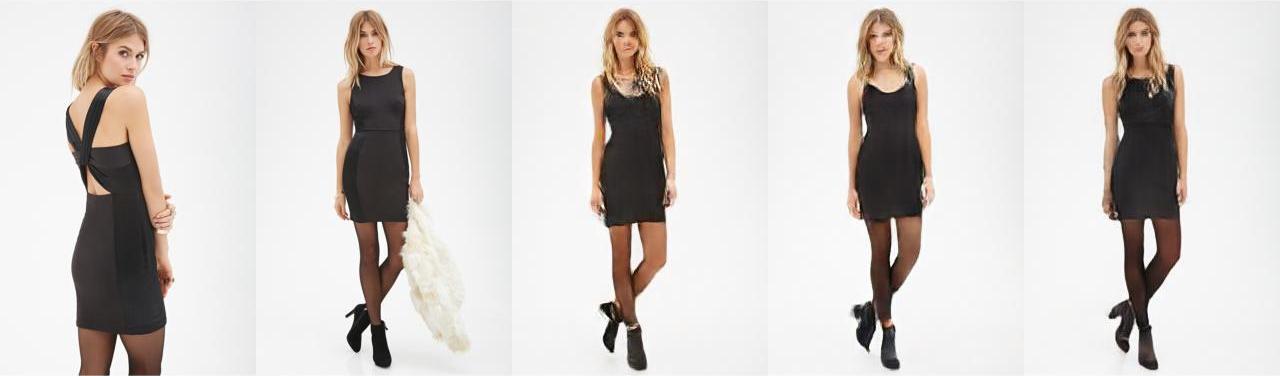}\\    
\end{tabular}}
\end{center}
\caption{Sample results on the DeepFashion dataset. From left to right: source image, target image, \methodN w/ fixed body displacement field, \methodN w/ learned displacement field, \methodN full. Notice the superior quality of the images synthesized with learned displacement field.}
\label{fig:ablation_results}
\end{figure}

\subsection{Human Synthesis}

\mypar{Datasets used.} We train our model on the DeepFashion (In-shop Clothes Retrieval Benchmark) \cite{liuLQWTcvpr16DeepFashion} dataset, which contains $52,712$ in-shop clothes images and $\approx200,000$ cross-pose/scale pairs.  We use the train/test split provided by \cite{siarohin2018deformable}, containing $101,268$ train and $8,616$ test pairs of the same person in two different poses.

\mypar{Quantitative Evaluation.} In table \ref{tab:eval_fasion} we provide quantitative results of \methodN on the DeepFashion dataset and compare them with previous works. In addition, we perform an ablation study to show the impact of the learned displacement field in the quality of the generated images. Our first baseline (\methodN w/o displacement field) is a simplified version that does not use a displacement field and its associated encoder. The source image, $I_s$, source dense pose $D_s$, source clothing segmentation $C_s$, and target dense pose $D_t$ are concatenated and fed directly to the \textit{Appearance Encoder}. As can be seen in table \ref{tab:eval_fasion}, this is a strong baseline, achieving competitive results with previous works under the IS metric, and state-of-the-art results when measuring SSIM. Its strength lies in leveraging more complex pose and shape information in the form of dense pose (previously only used by DPT \cite{neverova2018dense}). In the second baseline (\methodN w/ fixed body displacement field), we compute the body displacement field between the 3d body model of the source image and that of the target image, and use it to warp the encoded appearance features of the source. Even if this displacement field is fixed and limited only to body regions, it still improves the results over the first baseline. In the third baseline, we use a dedicated \textit{Topology Encoder} (see fig. \ref{fig:HS_overview}) to learn a correction of the initial body displacement, that is extended to clothes and hair. The appearance features of the source image are warped with the corrected motion field, right before the residual blocks computation. By doing so, we obtain the largest performance increase, as well as state-of-the-art results on both IS and SSIM scores. This shows the positive impact that the learned displacement field has on the quality of the generated images. The full method (\methodN full) adds two elements: i) the deformable skip connections between the encoder and the decoder for faster convergence and ii) the target dense pose in the decoding phase, such that information on the position in which the person should be synthesized is explicitly available.

\begin{table}[htb]
    \begin{center}
     \scalebox{0.80}{
        \begin{tabular}{|l|c|c|}
        \hline
        \textbf{Methods} & \textbf{IS \cite{salimans2016improved}} $\mathbf{\uparrow}$ & \textbf{SSIM \cite{Wang04imagequality}} $\mathbf{\uparrow}$ \\
        \hline
        \hline
        \textbf{pix2pix}\cite{isola2017} (CVPR17)& $3.249$ & $0.692$  \\
        \hline
        \textbf{PG2}\cite{ma2017pose} (NEURIPS17) & $ 3.090$ & 0.$762$  \\
        \hline
        \textbf{DSCF}\cite{siarohin2018deformable} (CVPR18) &  $3.351$ &  $0.756$   \\
        \hline
         \textbf{UPIS}\cite{UPIS} (CVPR18) & $2.970$ &  $0.747$  \\
        \hline
         \textbf{BodyROI7}\cite{ma2018disentangled} (CVPR18) & $3.228$ &  $0.614$   \\
        \hline
        \textbf{AUNET}\cite{Esser_2018_CVPR} (CVPR18) & $3.087$& $0.786$ \\ 
        \hline 
        \textbf{DPT}\cite{neverova2018dense} (ECCV18) & $3.61$ & $0.785$ \\
        \hline
        \textbf{SGW-GAN}\cite{Soft-Gated_nips18} (NEURIPS18) & $3.446$& $0.793$ \\
        \hline
        \textbf{DIAF} \cite{li2019dense} (CVPR19) & $3.338$ & $0.778$\\
        \hline
        \hline
        \textbf{\methodN w/o displacement field} & $ 3.3876 $ & $ 0.8035$   \\
        \hline
        \textbf{\methodN w/ fixed body displacement field} & $3.4541$ & $0.8016$ \\
        \hline
        \textbf{\methodN w/ learned displacement field } &  $3.6343$ & $0.8049$ \\
        \hline
        \textbf{\methodN full} &  $\mathbf{3.6950}$ & $\mathbf{0.8135}$ \\
        \hline
        \end{tabular}
        }
    \end{center}
    \caption{Evaluation on the DeepFashion dataset. We perform several ablation studies, showing that learning the displacement field, compared to no or fixed body displacement, performs better in terms of both IS and SSIM scores. Our full method obtains state-of-the art results when compared to previous works.  }
    \label{tab:eval_fasion}
\end{table}  

\begin{figure*}[!htbp]
\begin{center}
\scalebox{0.55}{
\setlength{\tabcolsep}{2pt}
\begin{tabular}{c|cccccccccc}
\includegraphics[height=108pt]{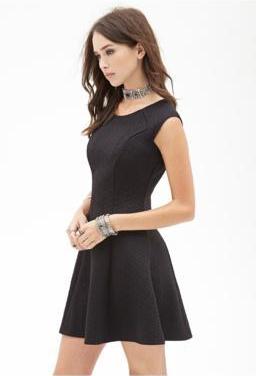} &
 \includegraphics[height=108pt]{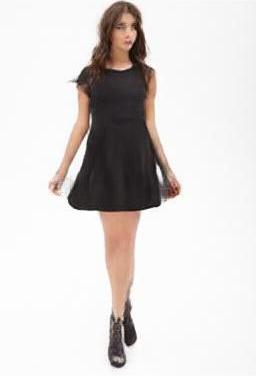} &
         \includegraphics[height=108pt]{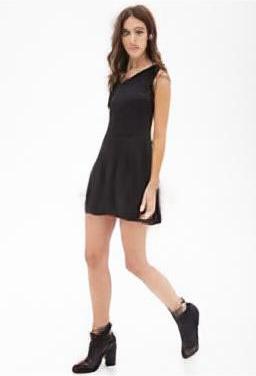} &
         \includegraphics[height=108pt]{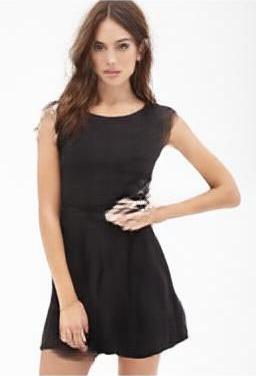} &
         \includegraphics[height=108pt]{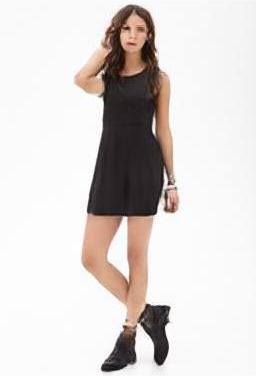} &
         \includegraphics[height=108pt]{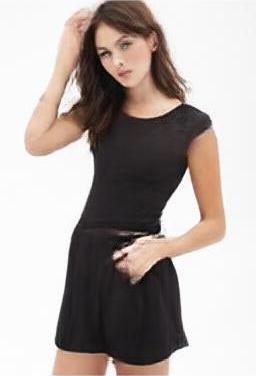} &
           
             \includegraphics[height=108pt]{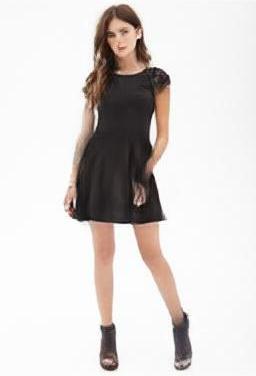} & 
                 \includegraphics[height=108pt]{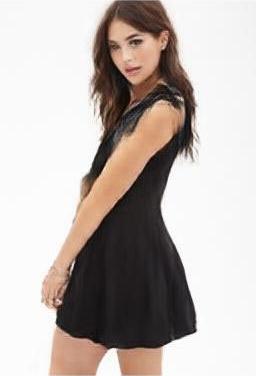} &
               \includegraphics[height=108pt]{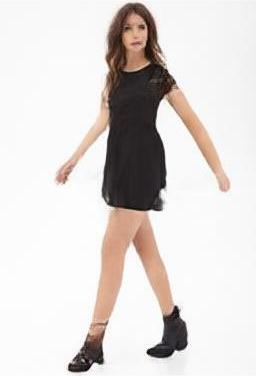} &
                   \includegraphics[height=108pt]{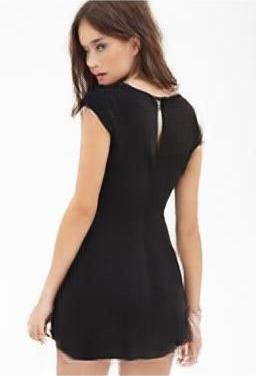} &
          \includegraphics[height=108pt]{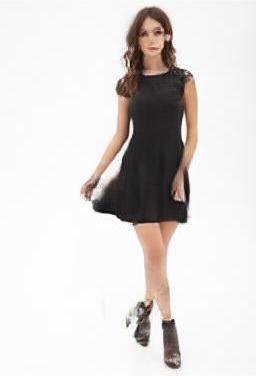}   \\
          \includegraphics[height=108pt]{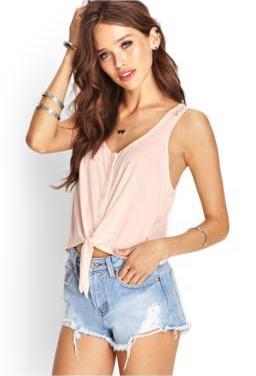} &
          \includegraphics[height=108pt]{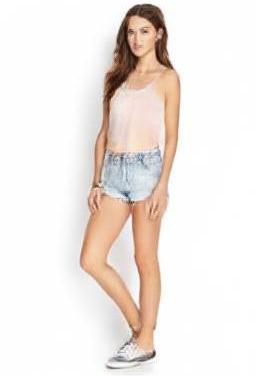} &
\includegraphics[height=108pt]{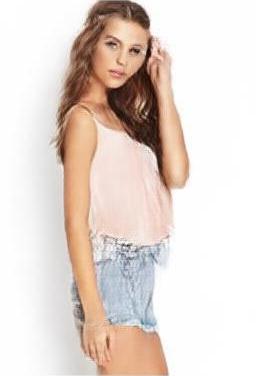} &
\includegraphics[height=108pt]{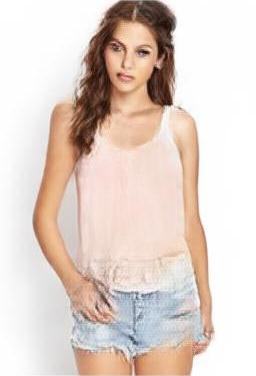} &
\includegraphics[height=108pt]{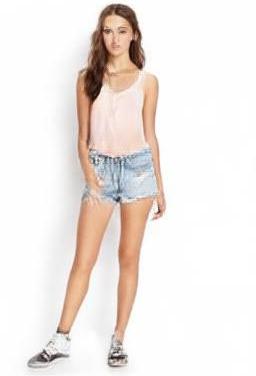} &
\includegraphics[height=108pt]{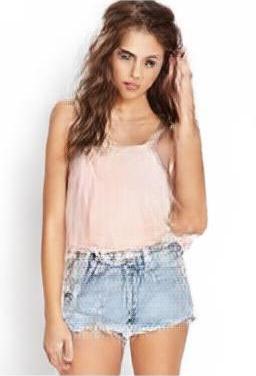} &
\includegraphics[height=108pt]{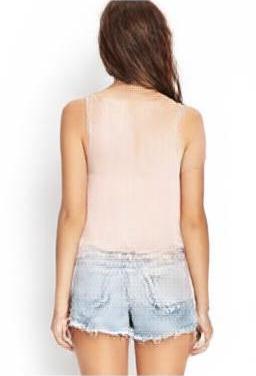} &
\includegraphics[height=108pt]{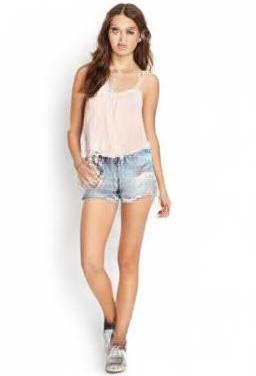} &
\includegraphics[height=108pt]{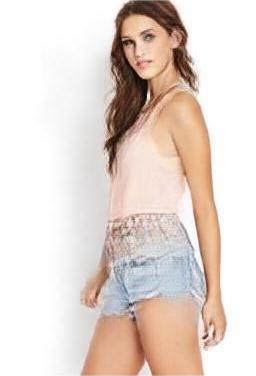} &
\includegraphics[height=108pt]{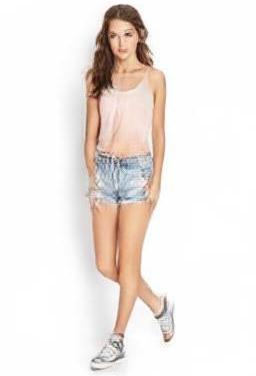} &
\includegraphics[height=108pt]{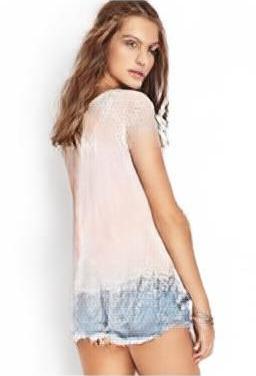}
  \end{tabular}
  }  
 
\end{center}
\caption{Appearance transfer results of a single RGB image into various poses. The first image of each row represents the source, while the others are obtained by synthesizing that person in a different pose. }
\label{fig:various_poses}
\end{figure*}

\begin{figure*}[!htbp]
\begin{center}

         \includegraphics[width=0.3\linewidth]{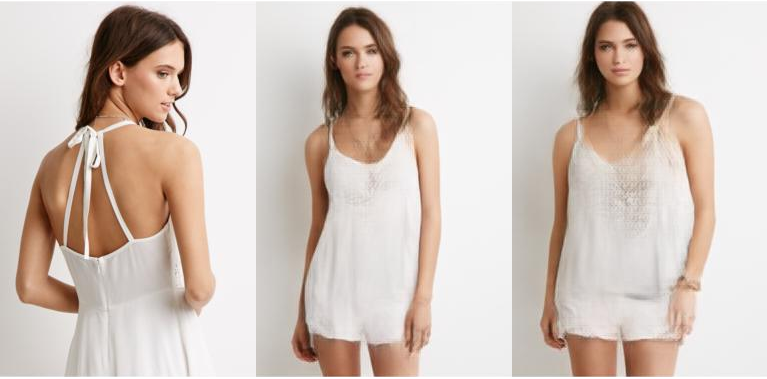} 
         \includegraphics[width=0.3\linewidth]{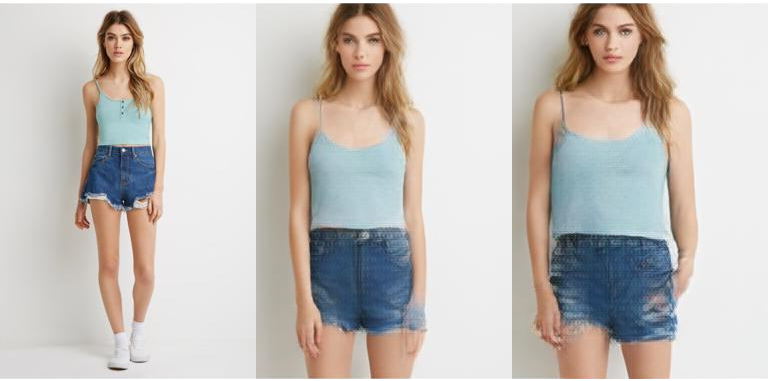}
         \includegraphics[width=0.3\linewidth]{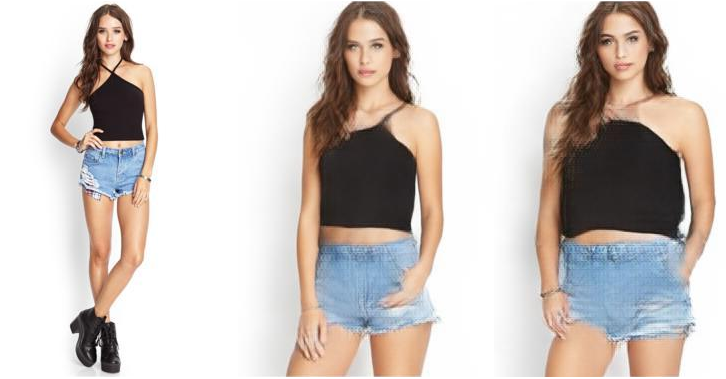}

\end{center}
\caption{Human synthesis with varying shape parameters. \textbf{(Left)} Source image. \textbf{(Center)} Synthesized image with same shape parameters as the source. \textbf{(Right)} Synthesized image with \textit{larger} shape parameters as the source.}
\label{fig:shape_result}
\end{figure*}

\begin{figure*}[!htbp]
\begin{center}
         \includegraphics[width=0.3\linewidth]{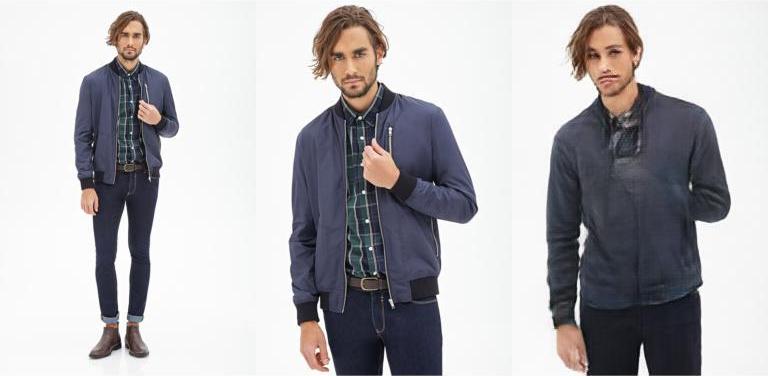} 
         \includegraphics[width=0.3\linewidth]{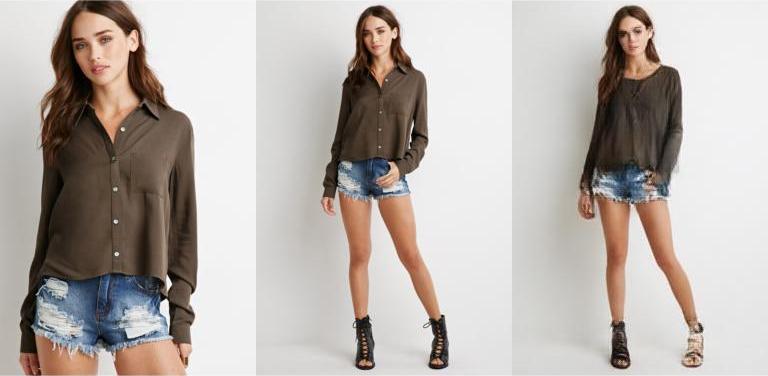}
         \includegraphics[width=0.3\linewidth]{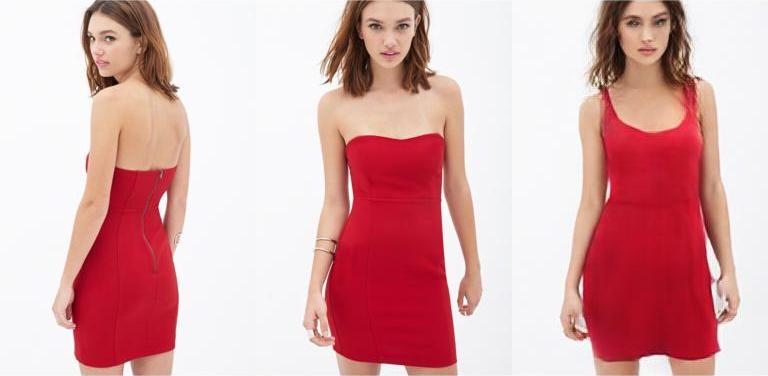}
    
\end{center}
\caption{Example of failure cases.  \textbf{(Left)} Source image.   \textbf{(Center)} Ground truth target image. \textbf{(Right)} Synthesized image.  When the body model of the target image is misfitted, small body parts such as hands may become blurry or disappear \textbf{(first example)}.  Clothing items that are not visible in the source image (shoes) can be generated inconsistently since we do not enforce a symmetry of the generated clothing \textbf{(second example)}. Clothing completion in the synthesized image may not respect the full level of detail in the source: the synthesized straps are not present in the source image \textbf{(third example)}. }
\label{fig:failure}
\end{figure*}

\mypar{Qualitative Evaluation.} A visual comparison of the baselines and the full method can be seen in fig. \ref{fig:ablation_results}. The first column represents the source image, the second column is the ground truth target image, while the last three columns are outputs of the following methods: \methodN w/ fixed body displacement field, \methodN w/ learned displacement field and \methodN full. Notice the superior quality of the images synthesized with learned displacement (columns 4 and 5) in terms of i) \textit{pose}: the order in which the legs overlap in the first image, ii) \textit{clothing details}: the folds of the dress in the fourth image look more realistic when using learned motion field, iii) \textit{color/texture preserving}: the color in the third row and the floral pattern in the fifth row better resemble those of source image and iv) \textit{face details}: in all cases the face of the synthesized person is sharper and better resembles the source person when using the full method. Fig. \ref{fig:ablation_results} shows that our method can synthesize high-quality images of persons in a wide range of poses and with a large variety of clothing types, both closely fitted to the body and loose. Moreover, \methodN  allows altering the underlying 3d pose and shape of the desired target person, while correspondingly adjusting their appearance. Please see fig. \ref{fig:shape_result} for examples of body proportions variations, and fig. \ref{fig:various_poses} for synthesized images of the same person in various poses. Examples of failure cases of our network are shown in fig. \ref{fig:failure}.

\begin{figure*}[!htbp]
\begin{center}
         \includegraphics[height=94pt]{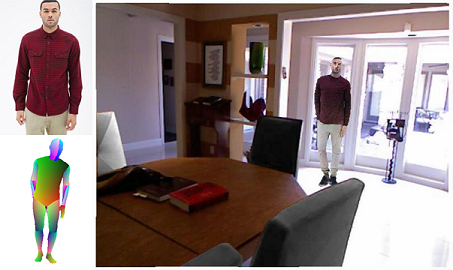}
         \includegraphics[height=94pt]{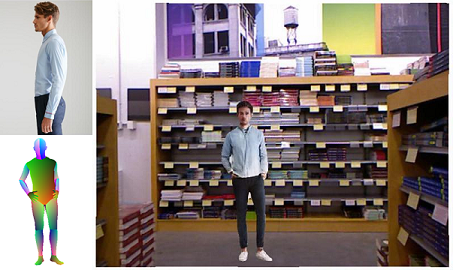}
         \includegraphics[height=94pt]{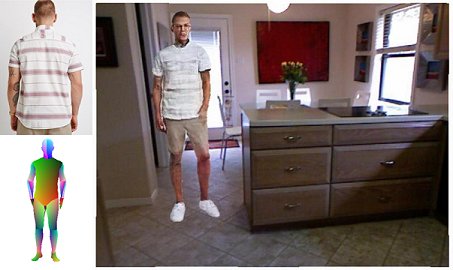}
         \\
        \includegraphics[height=94pt]{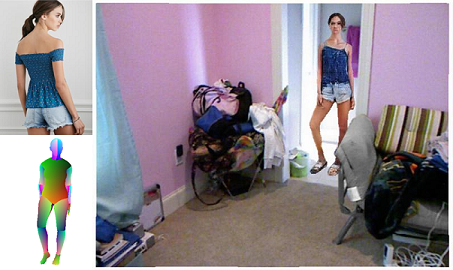} 
         \includegraphics[height=94pt]{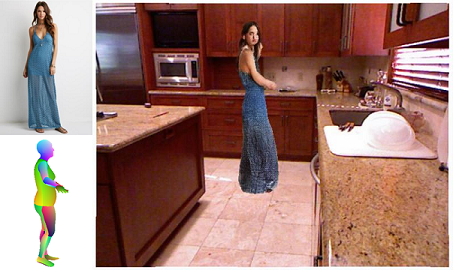}
         \includegraphics[height=94pt]{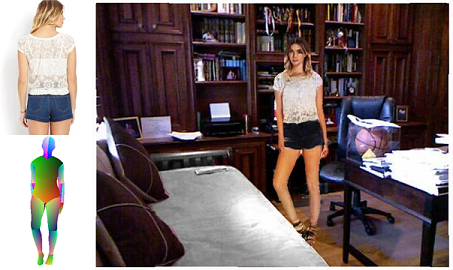}
         \\
         \includegraphics[height=78pt]{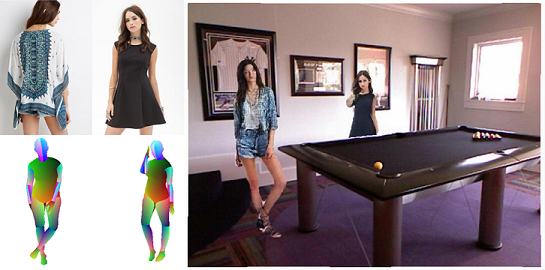}
         \includegraphics[height=78pt]{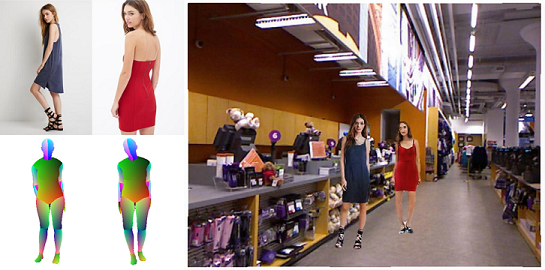}
         \includegraphics[height=78pt]{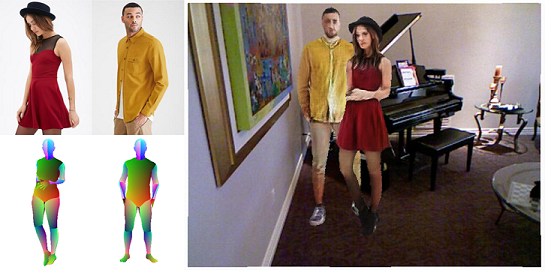}
\end{center}
\caption{Sample images generated by our proposed framework. For each example, we show the input source image, the target 3d body mesh, and a scene with a geometrically plausible placement of the synthesized person. Please note that our framework allows for a positioning behind various scene objects, and the insertion of multiple people without breaking any geometrical scene properties. }
\label{fig:ric_examples}
\end{figure*}

\subsection{Appearance and Geometrical Compositing}

\mypar{Datasets used.} For training the Appearance Compositing network, we use the COCO \cite{Lin14} dataset because of its large variety of natural images, containing both humans and associated ground truth segmentation masks. For the Geometric Compositing pipeline, we sample various backgrounds from the NYU Depth Dataset V2 \cite{Silberman:ECCV12}. This dataset contains $1,449$ RGB images, depicting $464$ indoor scenes captured with a Kinect sensor. For each image, the dataset provides corresponding semantic annotations as well as aligned depth images. The choice of this dataset is motivated by multiple factors: the variety of backgrounds it provides, and the annotated depth and semantic class labels, that can be used to infer physically plausible space configurations where a human model can be placed, while obeying per-pixel depth ordering. We consider the semantic classes "floor", "rug", "floor mat" and "yoga mat" for our support surface. 

\mypar{Quantitative Evaluation.} Table \ref{tab:LPIPS} shows the performance of our Appearance Compositing network on images from the COCO validation dataset under the LPIPS metric. We first show the LPIPS distance between the initial images and their randomly perturbed versions ($0.0971$). Then we show how this difference is considerably reduced when we apply our Appearance Compositing network, that uses VGG and L1 loss (from $0.0971$ to  $0.0588$). Note that there is also a considerable reduction in the standard deviation. The results further improve if we add a discriminator head to our network (from $0.0588$  to $0.0542$ ).

\begin{table}[htb]
    \begin{center}
     \scalebox{0.74}{
        \begin{tabular}{|l|c|c|c|}
        
        \hline &  \multicolumn{2}{|c|}{\textbf{LPIPS score \cite{zhang2018perceptual}}} \\ 
        \hline
        \textbf{Reference} & \textbf{Full image} $\mathbf{\downarrow}$ & \textbf{Foreground} $\mathbf{\downarrow}$\\
        \hline
        \hline
        Perturbed & $ 0.0971 \pm 0.0596$ & $ 0.2569 \pm 0.0811$ \\
        \hline
        Refined (L1+VGG loss) & $ 0.0588 \pm 0.0267$ & $ 0.1317 \pm 0.0514$ \\
        \hline
        \textbf{Refined (L1+VGG+GAN loss)} & {$ \mathbf{0.0542 \pm 0.0243}$} & $ \mathbf{0.1165 \pm 0.0480}$ \\
        \hline
        \end{tabular}
        }
    \end{center}
    \caption{Learned Perceptual Image Patch Similarity metric (smaller the better) on 1000 images from Microsoft COCO validation dataset.}
    \label{tab:LPIPS}
\end{table}

In table \ref{tab:inception_score} we show the Inception Score of $1000$ images generated by HUSC with humans on diverse backgrounds: with only the Geometrical Compositing, and when the Appearance Compositing module is also added. The score is improved when correcting the appearance of the geometrical composites, which indicate that the images become more realistic. As a comparison baseline, we checked the IS of the background images alone. Note that there is only a less than $10\%$ drop, in the score of the final composite image, as compared to the original (real), background image.

\begin{figure}[!htbp]
\begin{center}
        
\includegraphics[height=85pt]{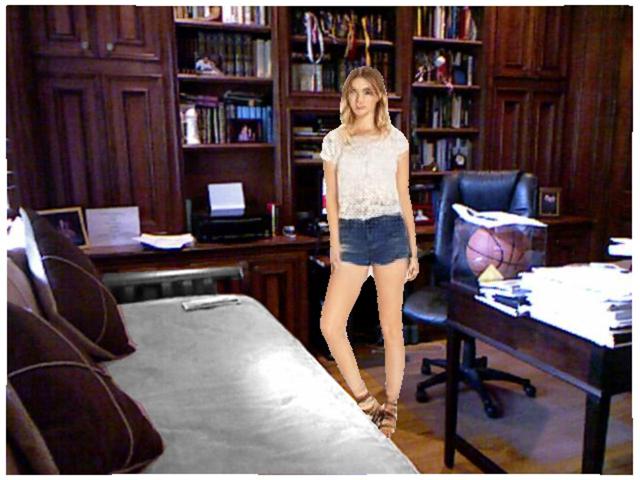}
\includegraphics[height=85pt]{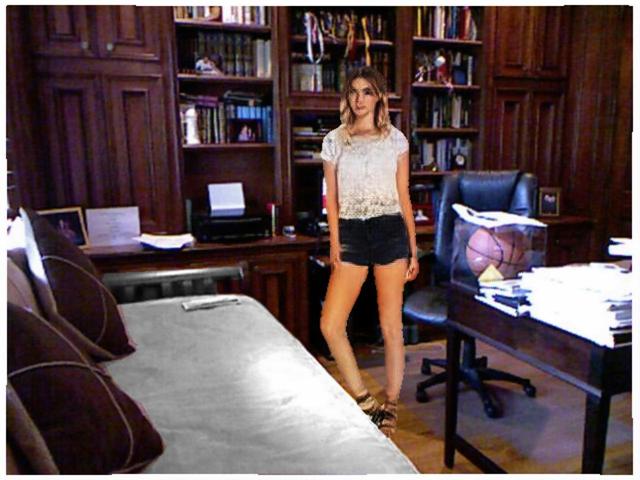}
\end{center}
\caption{ Before and after Appearance Compositing. We learn an adjustment of the color and boundaries of a human composite such that it blends naturally (in a statistical, learned sense) in the scene. 
}
\label{fig:results_coco}
\end{figure}

\begin{table}[!htb]
    \begin{center}
     \scalebox{0.80}{
        \begin{tabular}{|l|c|c|}
        \hline
        \textbf{Image set} & \textbf{Inception Score \cite{salimans2016improved}} $\mathbf{\uparrow}$\\
        \hline
        \hline
        \textbf{NYU  Depth Dataset V2} & $ 6.96 $  \\
        \hline
        \textbf{HUSC (Only Geometrical Compositing)} & $ 6.23$  \\
        \hline
        \textbf{HUSC } &  $6.33$  \\
        \hline
        \end{tabular}
        }
    \end{center}
    \caption{Inception Score (higher the better) for images generated by our method with and without appearance compositing. We also show the Inception Score for the original background images selected from the NYU dataset.}
    \label{tab:inception_score}
\end{table}  
\mypar{Qualitative Examples.} In fig. \ref{fig:ric_examples} we show examples of images generated by our proposed HUSC framework. For each example, we show the source image, the dense pose associated with the target body model and the resulting synthesized scene. Notice that the persons are naturally blended in the background scene at both geometrical and appearance levels.  In fig. \ref{fig:results_coco} 
we illustrate before and after results for our appearance compositing network. It can be noticed that that our method adapts the foreground to the ambient scene context. \textit{Please see the supplementary material.}

\section{Conclusions}
    We have presented a \textbf{HU}man \textbf{S}ynthesis and Scene \textbf{C}ompositing framework (\methodFull) for the realistic and controllable synthesis of humans with different appearance, in novel poses and scenes. By operating entirely in the 3d scene space rather than image space, except for late global image adaptation stages, and by taking into account scene semantics, we are able to realistically place the human impostor on support surfaces, handle scene scales and model occlusion. Moreover, by working with parametric 3d human models and dense geometric correspondences, we can better control and localize the appearance transfer process during synthesis. The model produces pleasing qualitative results and obtains superior quantitative results in the DeepFashion dataset, making it practically applicable for photo editing, fashion virtual try-on, or for the realistic data augmentation protocols used for training large scale 3d human sensing models.

    
{\small \noindent{\bf Acknowledgments:} This work was supported in part by the
European Research Council Consolidator grant SEED, CNCS-UEFISCDI (PN-III-P4-ID-PCE-2016-0535, PN-III-P4-ID-PCCF-2016-0180), the EU Horizon 2020 grant DE-ENIGMA (688835), and SSF.}
{\small
\bibliographystyle{ieee}
\bibliography{human}

\begin{thebibliography}{10}\itemsep=-1pt

\bibitem{alhaija20018geometric}
H.~A. Alhaija, S.~K. Mustikovela, A.~Geiger, and C.~Rother.
\newblock Geometric image synthesis.
\newblock {\em CoRR}, abs/1809.04696, 2018.

\bibitem{Güler_2018_CVPR}
R.~Alp~Güler, N.~Neverova, and I.~Kokkinos.
\newblock Densepose: Dense human pose estimation in the wild.
\newblock In {\em The IEEE Conference on Computer Vision and Pattern
  Recognition (CVPR)}, June 2018.

\bibitem{balakrishnan2018synthesizing}
G.~Balakrishnan, A.~Zhao, A.~V. Dalca, F.~Durand, and J.~Guttag.
\newblock Synthesizing images of humans in unseen poses.
\newblock In {\em CVPR}, 2018.

\bibitem{Soft-Gated_nips18}
H.~Dong, X.~Liang, K.~Gong, H.~Lai, J.~Zhu, and J.~Yin.
\newblock Soft-gated warping-gan for pose-guided person image synthesis.
\newblock In {\em Advances in Neural Information Processing Systems 31}, pages
  474--484. 2018.

\bibitem{dvornik2018modeling}
N.~Dvornik, J.~Mairal, and C.~Schmid.
\newblock Modeling visual context is key to augmenting object detection
  datasets.
\newblock In {\em Proceedings of the European Conference on Computer Vision
  (ECCV)}, pages 364--380, 2018.

\bibitem{dwibedi2017cut}
D.~Dwibedi, I.~Misra, and M.~Hebert.
\newblock Cut, paste and learn: Surprisingly easy synthesis for instance
  detection.
\newblock In {\em Proceedings of the IEEE International Conference on Computer
  Vision}, pages 1301--1310, 2017.

\bibitem{Esser_2018_CVPR}
P.~Esser, E.~Sutter, and B.~Ommer.
\newblock A variational u-net for conditional appearance and shape generation.
\newblock In {\em The IEEE Conference on Computer Vision and Pattern
  Recognition (CVPR)}, June 2018.

\bibitem{Gong_2018_ECCV}
K.~Gong, X.~Liang, Y.~Li, Y.~Chen, M.~Yang, and L.~Lin.
\newblock Instance-level human parsing via part grouping network.
\newblock In {\em The European Conference on Computer Vision (ECCV)}, September
  2018.

\bibitem{goodfellow2014}
I.~Goodfellow, J.~Pouget-Abadie, M.~Mirza, B.~Xu, D.~Warde-Farley, S.~Ozair,
  A.~Courville, and Y.~Bengio.
\newblock Generative adversarial nets.
\newblock In {\em NIPS}, 2014.

\bibitem{Grigorev2018CoordinatebasedTI}
A.~Grigorev, A.~Sevastopolsky, A.~T. Vakhitov, and V.~S. Lempitsky.
\newblock Coordinate-based texture inpainting for pose-guided image generation.
\newblock {\em CoRR}, abs/1811.11459, 2018.

\bibitem{han2018viton}
X.~Han, Z.~Wu, Z.~Wu, R.~Yu, and L.~S. Davis.
\newblock Viton: An image-based virtual try-on network.
\newblock In {\em CVPR}, 2018.

\bibitem{isola2017}
P.~Isola, J.-Y. Zhu, T.~Zhou, and A.~A. Efros.
\newblock Image-to-image translation with conditional adversarial networks.
\newblock In {\em CVPR}, 2017.

\bibitem{Johnson2016Perceptual}
J.~Johnson, A.~Alahi, and L.~Fei-Fei.
\newblock Perceptual losses for real-time style transfer and super-resolution.
\newblock In {\em European Conference on Computer Vision}, 2016.

\bibitem{lassnergenerative}
C.~Lassner, G.~Pons-Moll, and P.~V. Gehler.
\newblock A generative model of people in clothing.
\newblock In {\em ICCV}, volume~2, page~5, 2017.

\bibitem{DRIT}
H.-Y. Lee, H.-Y. Tseng, J.-B. Huang, M.~K. Singh, and M.-H. Yang.
\newblock Diverse image-to-image translation via disentangled representations.
\newblock In {\em ECCV}, 2018.

\bibitem{li2019dense}
Y.~Li, C.~Huang, and C.~C. Loy.
\newblock Dense intrinsic appearance flow for human pose transfer.
\newblock In {\em (CVPR)}, 2019.

\bibitem{Lin14}
T.-Y. Lin, M.~Maire, S.~Belongie, J.~Hays, P.~Perona, D.~Ramanan,
  P.~Doll{\'a}r, and C.~L. Zitnick.
\newblock Microsoft coco: Common objects in context.
\newblock In {\em ECCV}, pages 740--755. Springer, 2014.

\bibitem{liuLQWTcvpr16DeepFashion}
Z.~Liu, P.~Luo, S.~Qiu, X.~Wang, and X.~Tang.
\newblock Deepfashion: Powering robust clothes recognition and retrieval with
  rich annotations.
\newblock In {\em CVPR}, 2016.

\bibitem{SMPL2015}
M.~Loper, N.~Mahmood, J.~Romero, G.~Pons-Moll, and M.~J. Black.
\newblock {SMPL}: A skinned multi-person linear model.
\newblock {\em SIGGRAPH}, 34(6):248:1--16, 2015.

\bibitem{luan2017deep}
F.~Luan, S.~Paris, E.~Shechtman, and K.~Bala.
\newblock Deep photo style transfer.
\newblock In {\em CVPR}, volume~2, page~5, 2017.

\bibitem{ma2017pose}
L.~Ma, X.~Jia, Q.~Sun, B.~Schiele, T.~Tuytelaars, and L.~Van~Gool.
\newblock Pose guided person image generation.
\newblock In {\em NIPS}, pages 406--416, 2017.

\bibitem{ma2018disentangled}
L.~Ma, Q.~Sun, S.~Georgoulis, L.~Van~Gool, B.~Schiele, and M.~Fritz.
\newblock Disentangled person image generation.
\newblock In {\em CVPR}, pages 99--108, 2018.

\bibitem{Silberman:ECCV12}
P.~K. Nathan~Silberman, Derek~Hoiem and R.~Fergus.
\newblock Indoor segmentation and support inference from rgbd images.
\newblock In {\em ECCV}, 2012.

\bibitem{neverova2018dense}
N.~Neverova, R.~Alp~Guler, and I.~Kokkinos.
\newblock Dense pose transfer.
\newblock In {\em Proceedings of the European Conference on Computer Vision
  (ECCV)}, pages 123--138, 2018.

\bibitem{nguyen2016plug}
A.~Nguyen, J.~Yosinski, Y.~Bengio, A.~Dosovitskiy, and J.~Clune.
\newblock Plug \& play generative networks: Conditional iterative generation of
  images in latent space.
\newblock {\em arXiv preprint arXiv:1612.00005}, 2016.

\bibitem{UPIS}
A.~Pumarola, A.~Agudo, A.~Sanfeliu, and F.~Moreno{-}Noguer.
\newblock Unsupervised person image synthesis in arbitrary poses.
\newblock {\em The IEEE Conference on Computer Vision and Pattern Recognition
  (CVPR)}, 2018.

\bibitem{remez2018learning}
T.~Remez, J.~Huang, and M.~Brown.
\newblock Learning to segment via cut-and-paste.
\newblock In {\em Proceedings of the European Conference on Computer Vision
  (ECCV)}, pages 37--52, 2018.

\bibitem{salimans2016improved}
T.~Salimans, I.~Goodfellow, W.~Zaremba, V.~Cheung, A.~Radford, and X.~Chen.
\newblock Improved techniques for training gans.
\newblock In {\em NIPS}, pages 2234--2242, 2016.

\bibitem{siarohin2018deformable}
A.~Siarohin, E.~Sangineto, S.~Lathuili{\`e}re, and N.~Sebe.
\newblock Deformable gans for pose-based human image generation.
\newblock In {\em CVPR}, 2018.

\bibitem{tsai2017deep}
Y.-H. Tsai, X.~Shen, Z.~Lin, K.~Sunkavalli, X.~Lu, and M.-H. Yang.
\newblock Deep image harmonization.
\newblock In {\em CVPR}, volume~2, 2017.

\bibitem{varol17_surreal}
G.~Varol, J.~Romero, X.~Martin, N.~Mahmood, M.~J. Black, I.~Laptev, and
  C.~Schmid.
\newblock Learning from synthetic humans.
\newblock In {\em CVPR}, 2017.

\bibitem{wang2018pix2pixHD}
T.-C. Wang, M.-Y. Liu, J.-Y. Zhu, A.~Tao, J.~Kautz, and B.~Catanzaro.
\newblock High-resolution image synthesis and semantic manipulation with
  conditional gans.
\newblock In {\em Proceedings of the IEEE Conference on Computer Vision and
  Pattern Recognition}, 2018.

\bibitem{Wang04imagequality}
Z.~Wang, A.~C. Bovik, H.~R. Sheikh, and E.~P. Simoncelli.
\newblock Image quality assessment: From error measurement to structural
  similarity.
\newblock {\em IEEE TRANS. IMAGE PROCESSING}, 13:600--612, 2004.

\bibitem{yi2017dualgan}
Z.~Yi, H.~R. Zhang, P.~Tan, and M.~Gong.
\newblock Dualgan: Unsupervised dual learning for image-to-image translation.
\newblock In {\em ICCV}, pages 2868--2876, 2017.

\bibitem{zanfir17}
A.~Zanfir, E.~Marinoiu, and C.~Sminchisescu.
\newblock {Monocular 3D Pose and Shape Estimation of Multiple People in Natural
  Scenes -- The Importance of Multiple Scene Constraints}.
\newblock In {\em CVPR}, 2018.

\bibitem{zanfir_nips2018}
A.~Zanfir, E.~Marinoiu, M.~Zanfir, A.-I. Popa, and C.~Sminchisescu.
\newblock Deep network for the integrated 3d sensing of multiple people in
  natural images.
\newblock In {\em NIPS}, 2018.

\bibitem{zanfir18human}
M.~Zanfir, A.~I. Popa, and C.~Sminchisescu.
\newblock Human appearance transfer.
\newblock In {\em CVPR}, 2018.

\bibitem{zhang2018perceptual}
R.~Zhang, P.~Isola, A.~A. Efros, E.~Shechtman, and O.~Wang.
\newblock The unreasonable effectiveness of deep features as a perceptual
  metric.
\newblock In {\em CVPR}, 2018.

\bibitem{zhang2018residual}
Y.~Zhang, Y.~Tian, Y.~Kong, B.~Zhong, and Y.~Fu.
\newblock Residual dense network for image super-resolution.
\newblock In {\em CVPR}, 2018.

\end{thebibliography}
}
\end{document}